\documentclass[conference]{IEEEtran}
\usepackage{todonotes}

\usepackage{stmaryrd}
\usepackage{amsfonts}
\usepackage{graphicx,times,amsmath} % Add all your packages here
\usepackage{url}
\usepackage{epstopdf}
\IEEEoverridecommandlockouts    % to create the author's affliation portion
\usepackage[english]{babel}
\usepackage{microtype}

\begin{document}

% paper title: Must keep \ \\ \LARGE\bf in it to leave enough margin.
\title{\ \\ \LARGE\bf Anomaly Detection Based on Indicators
Aggregation\thanks{Tsirizo Rabenoro and J\'{e}r\^{o}me Lacaille are with the
  Health Monitoring Department, Snecma, Safran Group, Moissy Cramayel,
  France. (email: \{tsirizo.rabenoro, jerome.lacaille\}@snecma.fr).}
\thanks{Marie Cottrell and Fabrice Rossi are with the SAMM (EA 4543),
  Universit\'{e} Paris 1, Paris, France (email: \{marie.cottrell, fabrice.rossi\}@univ-paris1.fr)}}

\author{Tsirizo Rabenoro%
%\thanks{Hello. }%
\and J\'{e}r\^{o}me Lacaille \and Marie Cottrell\and Fabrice Rossi}

% avoiding spaces at the end of the author lines is not a problem with
% conference papers because we don't use \thanks or \IEEEmembership
% use only for invited papers
%\specialpapernotice{(Invited Paper)}

% make the title area
\maketitle

\begin{abstract}
  Automatic anomaly detection is a major issue in various areas. Beyond mere
  detection, the identification of the source of the problem that produced the
  anomaly is also essential. This is particularly the case in aircraft engine
  health monitoring where detecting early signs of failure (anomalies) and
  helping the engine owner to implement efficiently the adapted maintenance
  operations (fixing the source of the anomaly) are of crucial importance to
  reduce the costs attached to unscheduled maintenance. 

  This paper introduces a general methodology that aims at classifying
  monitoring signals into normal ones and several classes of abnormal
  ones. The main idea is to leverage expert knowledge by generating a very
  large number of binary indicators. Each indicator corresponds to a fully
  parametrized anomaly detector built from parametric anomaly scores designed
  by experts. A feature selection method is used to keep only the most
  discriminant indicators which are used at inputs of a Naive Bayes
  classifier. This give an interpretable classifier based on interpretable
  anomaly detectors whose parameters have been optimized indirectly by the
  selection process. The proposed methodology is evaluated on simulated data
  designed to reproduce some of the anomaly types observed in real world
  engines. 
\end{abstract}
%\todo[inline]{Regler l'affichage des keywords}
%  \keywords{ Health Monitoring; Turbofan; Fusion; Anomaly Detection }
% no key words

\section*{Acknowledgment}
This study is supported by grant from Snecma\footnote{Snecma, Safran Group, is one of the world’s leading manufacturers of aircraft and rocket engines, see
 \url{http://www.snecma.com/} for details.}.

\section{Introduction}\label{sec:intro}
\PARstart{A}{utomatic} anomaly detection is a major issue in numerous areas
and has generated a vast scientific literature
\cite{chandola2009anomaly}. Among the possible choices, statistical techniques
for anomaly detection are appealing because they can leverage expert knowledge
about the expected normal behavior of the studied system in order to
compensate for the limited availability of faulty observations (or more
generally of labelled observations). Those techniques are generally based on a
stationarity hypothesis: if for instance the studied system is monitored via a
series of real valued observations $X_1,\ldots,X_n$, then the $X_i$ are
assumed to be identically distributed under normal conditions. Detecting
an anomaly amounts to detecting a change in the probability distribution of
the $X_i$, at some point $k$, for example a change in the mean value from
$\mu_1$ for $X_1,\ldots,X_k$ to $\mu_2$ for $X_{k+1},\ldots,X_n$. Numerous
parametric and nonparametric methods have been proposed to achieve this goal
\cite{basseville1995detection}.

However, statistical tests efficiency is highly dependent to the adequacy
between the assumed data distribution and the actual data distribution. While
this is obvious for parametric tests, it also applies to non parametric ones
as, in general, they are not as efficient as parametric ones when the data
distribution is known. In addition, statistical methods rely on
meta-parameters, such as the length of the time window on which a change is
looked for, that have to be tuned to give maximal efficiency.

This article proposes to combine a (supervised) classification approach to
statistical techniques in order to obtain an automated anomaly detection
system that leverages both expert knowledge and labelled data sets. The main
idea consists in building from expert knowledge a large number of binary
indicators that correspond to anomaly detection decisions taken by statistical
tests suggested by the experts, with varying (meta)-parameters. Then a feature
selection method is applied to the high dimensional binary vectors to select
the most discriminative ones, using a labeled data set. Finally, a classifier
is trained on the reduced binary vectors to provide automatic detection for
future samples. 

This approach has numerous advantages over using classification or statistical
tests only. On the classification point of view, it has been shown in
e.g. \cite{fleuret-2004} that selecting relevant binary features among a large
number of simple features can lead to very high classification accuracy in
complex tasks. In addition, using features designed by experts allows one to
at least partially interpret the way the classifier is making decisions as
none of the features will be off a black box nature. This is particularly
important in our application context (see Section \ref{sec:context}). The
indicators play also a homogenisation role by hiding the complexity of the
signals in which anomalies are looked for (in a way similar to the one used in
\cite{hegedus2011methodology}, for instance). On the statistical point of
view, the proposed approach brings a form of automated tuning: a test
recommended by an expert can be included in numerous variants, with a
different set of meta-parameters per variant. The feature selection process
keep then the most adapted parameters.

\begin{figure}[htp]
\centering
\includegraphics[width=0.9\linewidth]{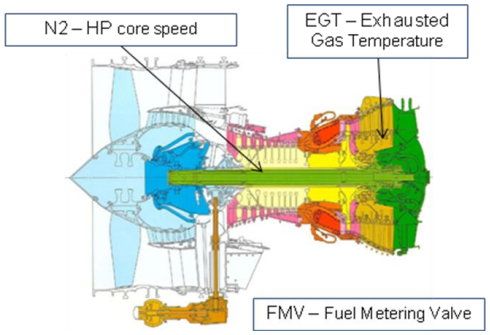}
\caption{Localization of some sensors embedded in an engine.}
\label{fig:engine}
\end{figure}

This methodology can be applied in various areas. This article focuses on
aircraft engine health monitoring which aims at detecting early signs of
failure to prevent from the occurrence of operational events (such as air turn
back). This detection is done through the analysis of data originating from
sensors embedded in the engine.  For example, messages on the Aircraft
Communications Addressing and Reporting System (ACARS\footnote{The ACARS is a
  standard system used to transmit messages between an aircraft and ground
  stations, see \url{http://en.wikipedia.org/wiki/ACARS} for details.}) give
an overview of engines status, and provide useful measurements at specific
moments that have been deemed important by experts. Flight after flight,
measurements, such as exhausted gas temperature (EGT) and high
pressure (HP) core speed (N2) (see Figure \ref{fig:engine}) form a time series
on which anomaly detection may be applied to detect early signs of failure. 

As aircraft engines are extremely reliable, labelled data including early
signs of failure are very scarce and not in a sufficient quantity to build
reliable fully automated detection systems. The methodology proposed in this
paper is therefore evaluated on simulated data in order to demonstrate its
efficiency and to justify the very costly collection of labelled data. 

The rest of the paper is organized as follows. Section \ref{sec:context}
describes in more details Snecma's engine health monitoring context which
motivates this study.  Section \ref{sec:method} presents in more details the
proposed methodology. Section \ref{sec:results} presents the results obtained
on simulated data. 

\section{Application context}\label{sec:context}
\subsection{Introduction and Objectives}
The very high reliability of aircraft engines is obtained by regular and
scheduled maintenance operations but also via engine health monitoring. This
process consists in ground based monitoring of numerous measurements made
on the engine and its environment during the aircraft operation. One of the
goal of this monitoring it to detect abnormal behavior of the engine that are
early signs of potential failures. 

On the one hand, missing such an early sign can lead to operational events
such as air turn back and delay and cancellation. Such operational events can
cause customers’ disturbance but also higher maintenance costs. On the other
hand, a false alarm (detecting an anomaly when the engine is behaving
normally) can have costly consequences from a useless inspection operation to
a useless engine removal procedure. This has a high cost both money wise and
in terms of customers' disturbance.

Thus to minimize false alarm, each potential anomaly is analyzed by human
operators. They are in charge of confirming the anomaly and in identifying its
probable origin. This latter part allows to estimate the repair costs (when
needed) and/or the immobilization time. (Note that human operators submit
their recommendations to the company owing the engine.) 

The long term goal of engine manufacturers is to help companies to minimize
their maintenance costs by giving maintenance recommendations as accurate as
possible. This means improving the detection performances of early signs of
failure. However, the context makes this goal more difficult to achieve than
in other situations because of two factors. Firstly, human operators have a
very important role in the current industrial process: the goal is to help
them reach improved decisions thanks to a grey box classifier, mainly because
the complexity of the problem seems to prevent any fully automated decision
making. Secondly, the reliability of current engines makes very scarce data
that display abnormal behavior. In practice, the scheduled maintenance tends
to prevent early signs of anomaly to manifest. In addition, the labelling of
abnormal data has to be done by experts, which makes it very expensive
(especially considering the scarceness just mentioned). 

The methodology proposed in this paper aims at addressing the first factor by
leveraging expert knowledge and relying on feature selection to keep only a
small number of binary indicators. In order to justify the costs of collecting
a large set of labelled data, and thus to address the second factor, the
methodology is evaluated on artificial data.

\subsection{Health monitoring}
As mentioned in the Introduction, aircraft engines are equipped with multiple
sensors which measure several physical quantities such as the oil pressure,
high pressure and low pressure core speed, air temperature, oil temperature,
etc. (See Figure \ref{fig:engine}.) Engine health monitoring is mainly based
on such flight data.

Monitoring is strongly based on experts knowledge and field experience. 
Faults and early signs of failures are identified from suitable measurements 
associated to adapted computational transformations of the data. We refer the
reader to e.g. \cite{rabenoroinstants} for examples of the types of
measurements and transformations that can be used in practice. 

\begin{figure}[htp]
\centering
\includegraphics[width=\linewidth]{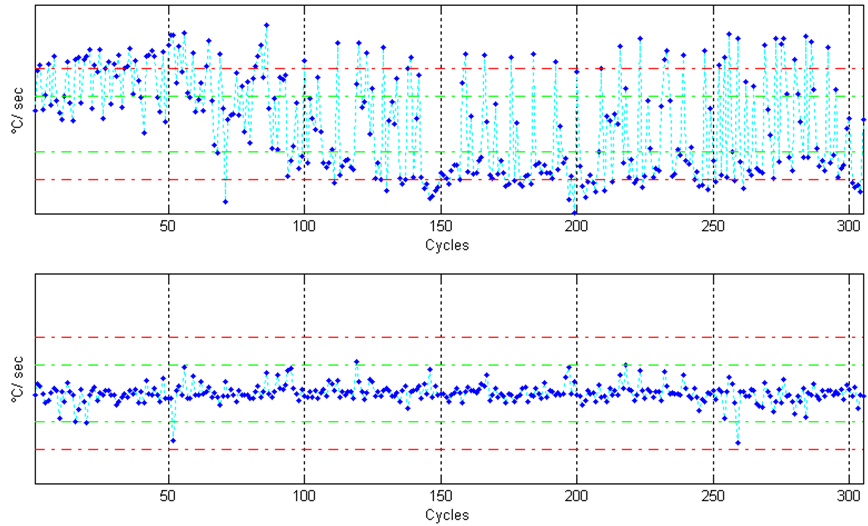}
\caption{Examples of results after preprocessing computation use to remove flight context dependency.}
\label{fig:norm}
\end{figure}

One of the main difficulty faced by the experts consists in removing from the
measurements any dependency from the flight context. (See Figure
\ref{fig:norm} for an example of such a transformation.) This normalization
process is extremely important as it allows one to assume stationarity of the
residual signal and therefore to leverage change detection methods. In
practice, experts build some anomaly score from those stationarity hypothesis
and when the score passes a limit, the corresponding early sign of failure is
signalled to the human operator. See \cite{come2010aircraft},
\cite{flandrois2009expertise} and \cite{lacaille2009maturation} for some
examples.

One of the problems induced by this general approach is that experts are
generally specialized on a particular subsystem, thus each anomaly score is
mainly focused on a particular subsystem despite the need of a diagnostic of
the whole system. This task is done by human operator who collects all
available information about the desired engine. One of the benefits of the
proposed methodology is its ability to handle binary indicators coming from
all subsystems in an integrated way, as explained in the next section.

\section{Methodology}\label{sec:method}
The proposed methodology aims at combining expert knowledge to supervised
classification in order to provide accurate and interpretable automatic
anomaly detection in the context of complex system monitoring. It is based on
the selection and combination of a large number of binary indicators. While
this idea is not entirely true (see e.g.,
\cite{fleuret-2004,hegedus2011methodology}), the methodology proposed here has
some specific aspects. Rather than relying on very basic detectors as in
\cite{fleuret-2004} or on fixed high level expertly designed ones as in
\cite{hegedus2011methodology}, our method takes an intermediate approach: it
varies the parameters of a set of expertly designed parametric indicators. In
addition, it aims at providing an interpretable model. This section details the
proposed procedure.

\subsection{Expert knowledge}\label{sec:expert-knowledge}
\begin{figure}
\centering
\includegraphics[width=0.75\linewidth]{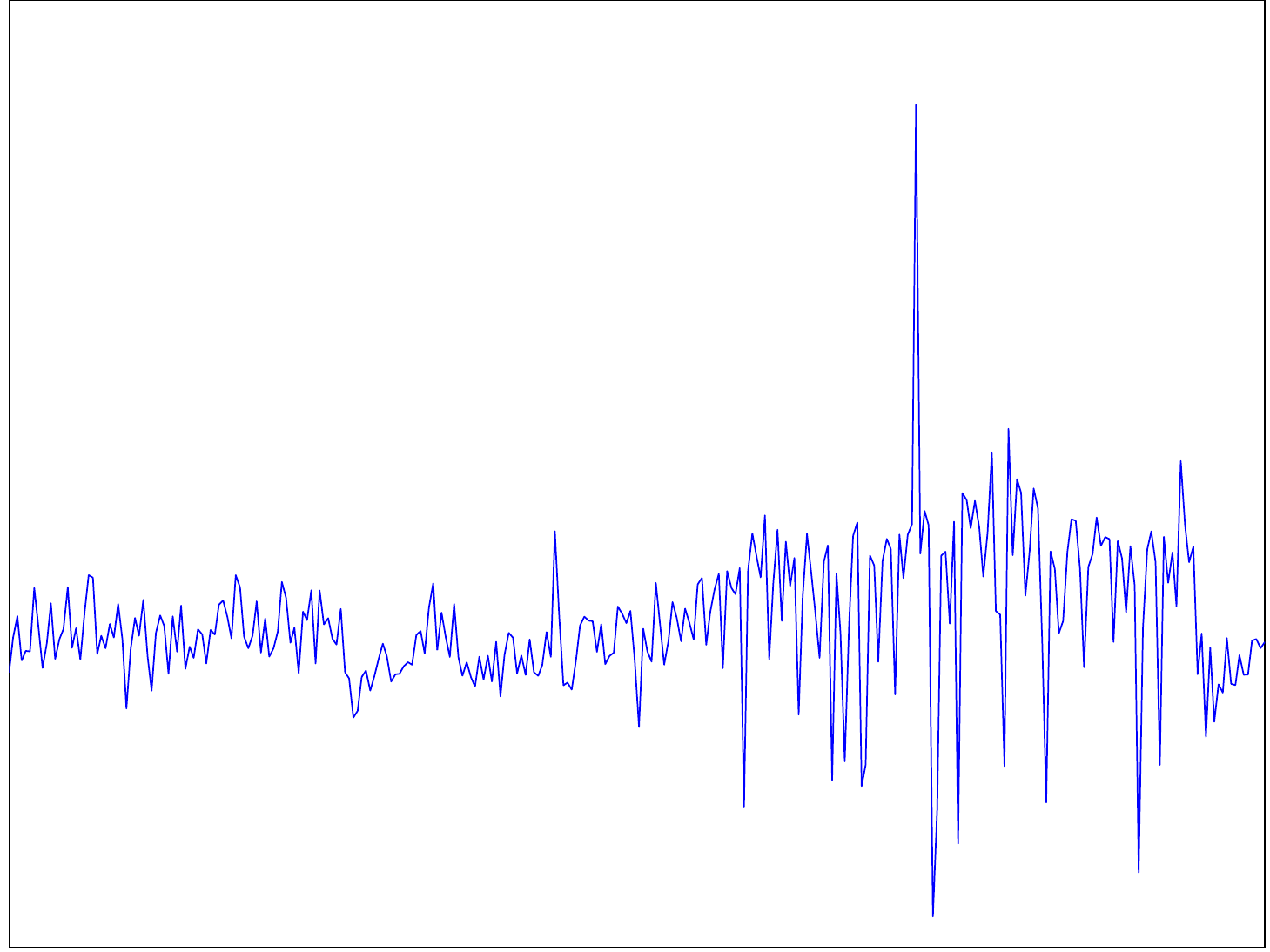}
\caption{Variance shift in a real world time series extracted from ACARS messages.}
\label{fig:varacars}
\end{figure}

\begin{figure}
\centering
\includegraphics[width=0.75\linewidth]{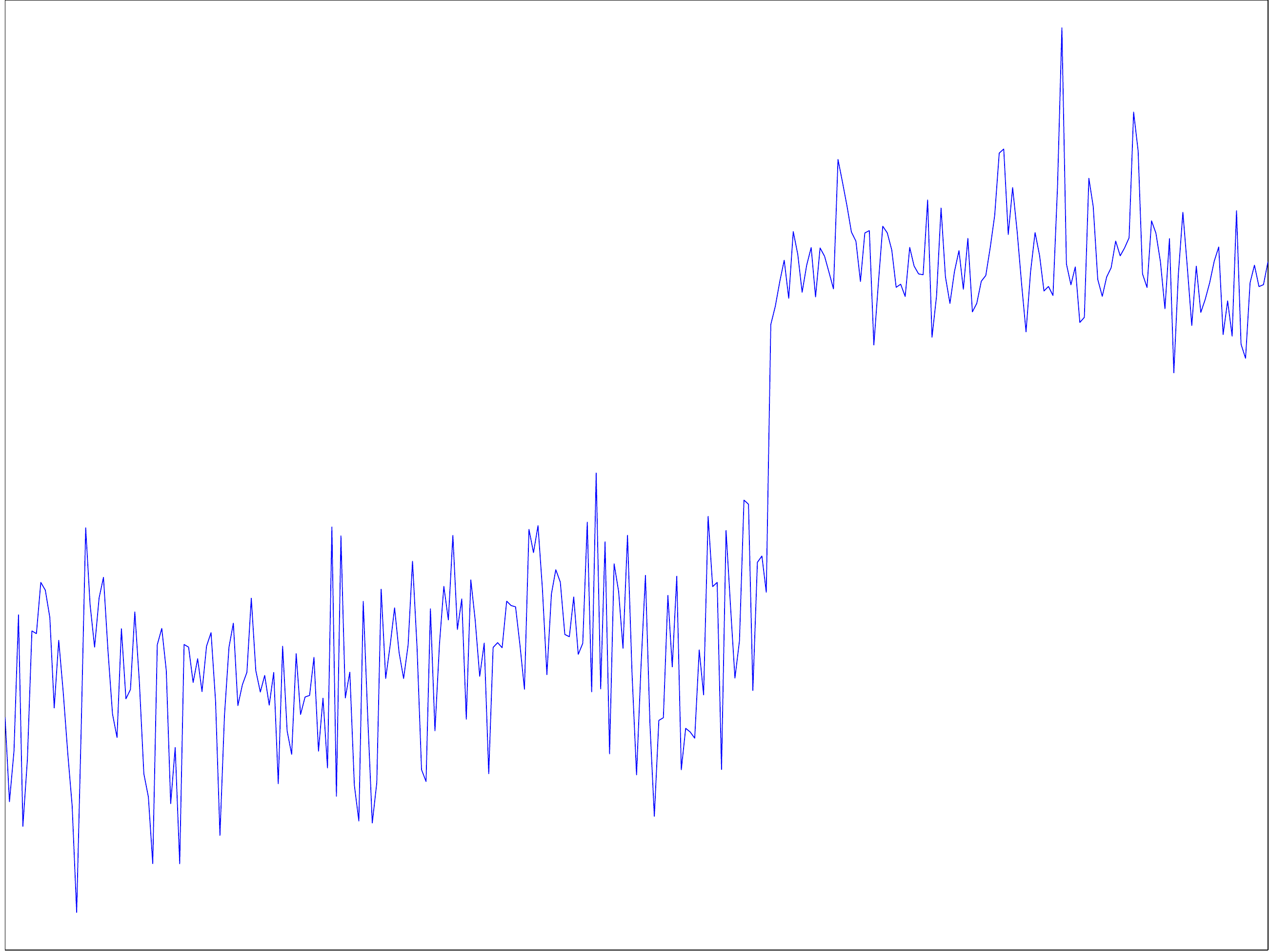}
\caption{Mean shift in a real world time series extracted from ACARS messages.}
\label{fig:meanacars}
\end{figure}

As explained in the introduction, this article focuses on change detection
based on statistical techniques \cite{basseville1995detection}. In many
contexts, experts can generally describe more or less explicitly the type of
change they are expecting for some specific (early signs of) anomalies. In the
proposed application context, one can observe for instance mean shift as in
Figure \ref{fig:meanacars} or variance shift as in Figure \ref{fig:varacars}. 


More generally, experts can described aggregation and transformation
techniques of raw signals that lead to quantities which should behave in a
``reasonable manner'' under normal circumstances. This can in general be
summarized by computing a distance between the actual quantities and there
expected values. 

\subsection{Exploring parameter space}
In practice however, experts can seldom provide detailed parameter settings
for the aggregation and transformation techniques they recommend. Fixing the
threshold above which a distance from the ``reasonable values'' becomes
critical is also difficult. 

Let us consider for illustration purpose that the expert recommends to look
for shifts in mean of a certain quantity as early signs of a specific anomaly
(as in Figure \ref{fig:meanacars}). If the expert believes the quantity to be
normally distributed with a fixed variance, then a natural test would be
Student's t-test. If the expert has no strong priors on the distribution, a
natural test would be the Mann–Whitney U test. 

Then, in both cases, one has to assess the scale of the shift. Indeed, those
tests work by comparing summary statistics of two populations, before and
after a possible change point. To define the populations, the expert has to
specify the length of the time windows to consider before and after the
possible change point: this is the expected scale at which the shift will
appear. In most cases, the experts can only give a rough idea of the
scale. 

Given the choice of the test, of its scale and of a change point, one can
construct a statistic, whose value can be turned into a $p$-value based on its
distribution under the null hypothesis (which would be stationarity in this
case). To take a decision, one has to choose a level to which the $p$-value
will be compared. 

So all in one, looking for a shift in mean can be done by choosing at least
three parameters: the type of the test, the scale at which the shift can occur
and the level of the test. For all these parameters, experts can give only
rough guidelines, in general. The proposed methodology consists in
considering (a subset of) all possible combinations of parameters compatible
with expert knowledge to generate binary indicators. In the present example,
this means choosing a finite set of scales and a finite set of levels, and
computing the decision of the tests obtained by applying both solutions
(t-test and U test) for all the combinations of levels and scales. This is a
form of indirect grid search procedures for meta-parameter optimisation. 

\subsection{Confirmatory indicators}\label{sec:conf-indic}
Finally, as pointed out before, aircraft engines are extremely reliable, a
fact that increases the difficulty in balancing sensibility and specificity of
anomaly detectors. In order to alleviate this difficulty, high level
confirmatory indicators are built from low level tests. For instance, if we
monitor the evolution of a quantity on a long period compared to the expected
time scale of anomalies, we can compare the number of times the null
hypothesis of a test has been rejected on the long period with the number of
times it was not rejected, and turn this into a binary indicator with a
majority rule.

\subsection{Decision}\label{sec:decision}
To summarize, we construct parametric anomaly scores from expert knowledge,
together with acceptable parameter ranges. By exploring those ranges, we
generate numerous (possible hundreds of) binary indicators. Each indicator can be
linked to an expertly designed score with a specific set of parameters and
thus is supposedly easy to interpret by operators. Notice that while we 
focused in this presentation on temporal data, this framework can be applied
to any data source. 
 
The final decision step consists in classifying these high dimensional binary
vectors in order to further discriminate between seriousness of anomalies
and/or sources (in terms of subsystems of the engine, for instance). For this,
a labelled data set is obviously needed. 

In the considered context, black box modelling is not acceptable, so while
numerous classification algorithms are available (see
e.g. \cite{kotsiantis2007supervised}), we shall focus on interpretable
ones. Random Forests \cite{breiman2001random} are chosen as the reference
method as they are very adapted to binary indicators and to high dimensional
data. They are also known to be robust and to provide state-of-the-art
classification performances at a very small computational cost. While they are
not as interpretable as their ancestors CART \cite{breiman1984classification},
they provide at least variable importance measures that can be used to
identify the most important indicators.

Another classification algorithms used in this paper is naive Bayes classifier
\cite{koller2009probabilistic} which is also appropriate for high dimensional
data.  They are known to provide good results despite the strong assumption of
the independence of features given the class. In addition, decisions taken by
a naive Bayes classifier are very ease to understand thanks to the estimation
of the conditional probabilities of the feature in each class. Those
quantities can be shown to the human operator as references. 

Finally, while including hundreds of indicators is important to give a broad
coverage of the parameter spaces of the expert scores and thus to maximize the
probability of detecting anomalies, it seems obvious that some redundancy will
appear. Unlike \cite{hegedus2011methodology} who choose features by random
projection, the proposed methodology favors interpretable solutions, even at
the expense of the classification accuracy: the goal is to help the human
operator, not to replace her/him. Thus feature selection
\cite{guyon2003introduction} is more appropriate. The reduction of number of
features will ease the interpretation by limiting the quantity of information
transmitted to the operators in case of a detection by the classifier. Among
the possible solutions, we choose to use the Mutual information based
technique Minimum Redundancy Maximum Relevance (mRMR, \cite{peng2005feature})
which was reported to give excellent results on high dimensional data (see
also \cite{fleuret-2004} for another possible choice).

\section{Experiments}\label{sec:results}
As pointed out in the introduction, labelling a sufficiently large data set in
the context of engine health monitoring will be a very costly task, mainly
because of the strong reliability of those engines. The proposed methodology
is therefore evaluated on simulated data which have been modelled based on
real world data such as the ones shown on Figures \ref{fig:varacars} and
\ref{fig:meanacars}. 

\subsection{Simulated data}
We consider univariate time series of variable length in which three types of
shifts can happen: the mean and variance shifts described in Section
\ref{sec:expert-knowledge}, together with a trend shift described below. Two
data sets are generated, $A$ and $B$. 

In both cases, it is assumed that expert based normalization has been
performed. Therefore when no shift in the data distribution occurs, we observe
a stationary random noise modeled by the standard Gaussian distribution, that
is $n$ random variables $X_1,\ldots, X_n$ independent and identically
distributed according to $\mathcal{N}(\mu=0,\sigma^2=1)$. Signals have a
length chosen uniformly at random between 100 and 200 observations 
(each signal has a specific length).

The three types of shift are:
\begin{enumerate}
\item a variance shift: in this case,
  observations are distributed according to
  $\mathcal{N}(\mu=0,\sigma^2)$ with $\sigma^2=1$ before the change point and $\sigma$ chosen uniformly
  at random in $[1.01, 5]$ after the change point;
\item a mean shift: in this case,
  observations are distributed according to
  $\mathcal{N}(\mu,\sigma^2=1)$ with $\mu=0$ before the change point and $\mu$ chosen uniformly
  at random in $[1.01, 5]$ after the change point in set $A$. Set $B$ is more
  difficult on this aspect as $\mu$ after the change point is chosen uniformly
  at random in $[0.505, 2.5]$;
\item a trend shift: in this case, observations are distributed according to
  $\mathcal{N}(\mu,\sigma^2=1)$ with $\mu=0$ before the change point and $\mu$
  increasing linearly from $0$ from the change point with a slope of chosen uniformly
  at random in $[0.02,3]$.
\end{enumerate}
Assume that the signal contains $n$ observations, then the change point is
chosen uniformly at random between the $\frac{2n}{10}$-th observation and the
$\frac{8n}{10}$-th observation. 

We generate according to this procedure two balanced data set with 6000
observations corresponding to 3000 observations with no anomaly, and 1000
observations for each of the three types of anomalies. The only difference
between data set $A$ and data set $B$ is the amplitude of the mean shift which
is smaller in $B$, making the classification harder.

\subsection{Indicators}
As explained in Section \ref{sec:method}, binary indicators are
constructed from expert knowledge by varying parameters, including scale and
position parameters. In the present context, sliding windows are used: for each
position of the window, a classical statistical test is conducted to decide
whether a shift in the signal occurs at the center of the window. 

The ``expert'' designed tests are here:
\begin{enumerate}
\item the Mann–Whitney–Wilcoxon U test (non parametric test for shift in
  mean);
\item the two sample Kolmogorov-Smirnov test (non parametric test for
  differences in distributions);
\item the F-test for equality of variance (parametric test based on a Gaussian
  hypothesis). 
\end{enumerate}
The direct parameters of those tests are the size of the window which defines
the two samples (30, 50, and $\min(n-2,100)$ where $n$ is the signal length) and
the level of significance of the test (0.005, 0.1 and 0.5). Notice that those
tests do not include a slope shift detection. 

Then, confirmatory indicators are generated, as explained in Section
\ref{sec:conf-indic}:
\begin{enumerate}
\item for each underlying test, the derived binary indicator takes the value
  one if on $\beta\times m$ windows out of $m$, the test detects a change. 
  Parameters are the test itself with its parameters, the value of $\beta$
  (we considered 0.1, 0.3 and 0.5) and the number of observations in common
  between two consecutive windows (the length of the window minus 1, 5 or
  10);
\item for each underlying test, the derived binary indicator takes the value
  one if on $\beta\times m$ consecutive windows out of $m$, the test detects a
  change (same parameters);
\item for each underlying test, the derived binary indicator takes the value
  one if there are 5 consecutive windows such that the test detects a change
  on at least $k$ of these 5 consecutive windows (similar parameters where $\beta$ is replaced by $k$).
\end{enumerate}
In addition, based on expert recommendation, all those indicators are applied
both to the original signal and to a smoothed signal (using a simple moving
average of 5 observations). 

More than 50 different configurations are used for each indicator, leading to a
total number of 810 binary indicators (it should be noted that only a subset
of all possible configurations is included into this indicator vector). 

\subsection{Performance analysis}
Each data set is split in a balanced way into a learning set with 1000 signals
and a test set with 5000 signals. We report the global classification accuracy
(the classification accuracy is the percentage of correct predictions,
regardless of the class) on the learning set to monitor possible over
fitting. The performances of the methodology are evaluated on 10 balanced
subsets of size 500 from the 5000 signals' test set. This allows to evaluate
both the average performances and their variability. For the Random Forest, we
also report the out-of-bag (oob) estimate of the classification accuracy (this
is a byproduct of the bootstrap procedure used to construct the forest, see
\cite{breiman2001random}). Finally, we use confusion matrices and class
specific accuracy to gain more insights on the results when needed. 

\subsection{Performances with all indicators}
As indicators are expertly designed and should cover the useful
parameter range of the tests, it is assumed that the best classification
performances should be obtained when using all of them, up to the effects of
the curse of dimensionality. 

\begin{table}
  \centering
  \begin{tabular}{lccc}
Data set & Training set acc. & OOB acc. & Test set average acc. \\\hline
$A$ & 0.9770 & 0.9228 &0.9352 (0.0100) \\
$B$ &  0.9709 & 0.9118 & 0.9226 (0.0108) \\\hline
  \end{tabular}
\medskip
  \caption{Classification accuracy of the Random Forest using the 810 binary
    indicators. For the test set, we report the average classification
    accuracy and its standard deviation between parenthesis.}
  \label{tab:fullRF}
\end{table}

Table \ref{tab:fullRF} reports the global classification accuracy of the
Random Forest, using all the indicators. As expected, Random Forests suffer
neither from the curse of dimensionality nor from strong over fitting (the
test set performances are close to the learning set ones). Table
\ref{tab:fullNBN} reports the same performance indicator for the Naive Bayes
classifier. Those performances are significantly lower than the one obtained
by the Random Forest. As shown by the confusion matrix on Table
\ref{tab:conf:NBNsetAtest}, the classification errors are not concentrated on
one class (even if the errors are not perfectly balanced). This tends to
confirm that the indicators are adequate to the task (this was already obvious
from the Random Forest).

\begin{table}[htbp]
  \centering
  \begin{tabular}{lccc}
Data set & Training set accuracy & Test set average accuracy \\\hline
$A$ &0.7856   & 0.7718 (0.0173) \\
$B$ & 0.7545 & 0.7381 (0.0178) \\\hline
  \end{tabular}
\medskip
  \caption{Classification accuracy of the Naive Bayes classifier using the 810 binary
    indicators. For the test set, we report the average classification
    accuracy and its standard deviation between parenthesis.}
  \label{tab:fullNBN}
\end{table}

\begin{table}
\centering
\begin{tabular}{c|ccccc}

&0&1&2&3&total\\\hline

0&1759&667&45&29&2500\\

1&64&712&50&3&829\\

2&7&2&783&37&829\\

3&32&7&195&595&829\\\hline

\end{tabular}
\caption{Data set $A$: confusion matrix with all indicators for Naive Bayes
  classifier on the full test set.}
\label{tab:conf:NBNsetAtest}
\end{table}

\subsection{Feature selection}
While the Random Forest give very satisfactory results, it would be
unacceptable for human operators as it operates in a black box way. While the
indicators have simple interpretation, it would be unrealistic to ask to an
operator to review 810 binary values to understand why the classifier favors
one class other the others. In addition, the performances of the Naive Bayes
classifier are significantly lower than those of the Random Forest one. Both
drawbacks favor the use of a feature selection procedure. 

As explained in Section \ref{sec:decision}, the feature selection relies on
the mRMR ranking procedure. A forward approach is used to evaluate how many
indicators are needed to achieve acceptable predictive performances. Notice
that in the forward approach, indicators are added in the order given by mRMR
and then never removed. As mRMR takes into account redundancy between the
indicators, this should not be a major issue. Then for each number of
indicators, a Random Forest and a Naive Bayes classifier are constructed and
evaluated. 

\begin{figure}[htbp]
\centerline{\includegraphics[width=\linewidth]{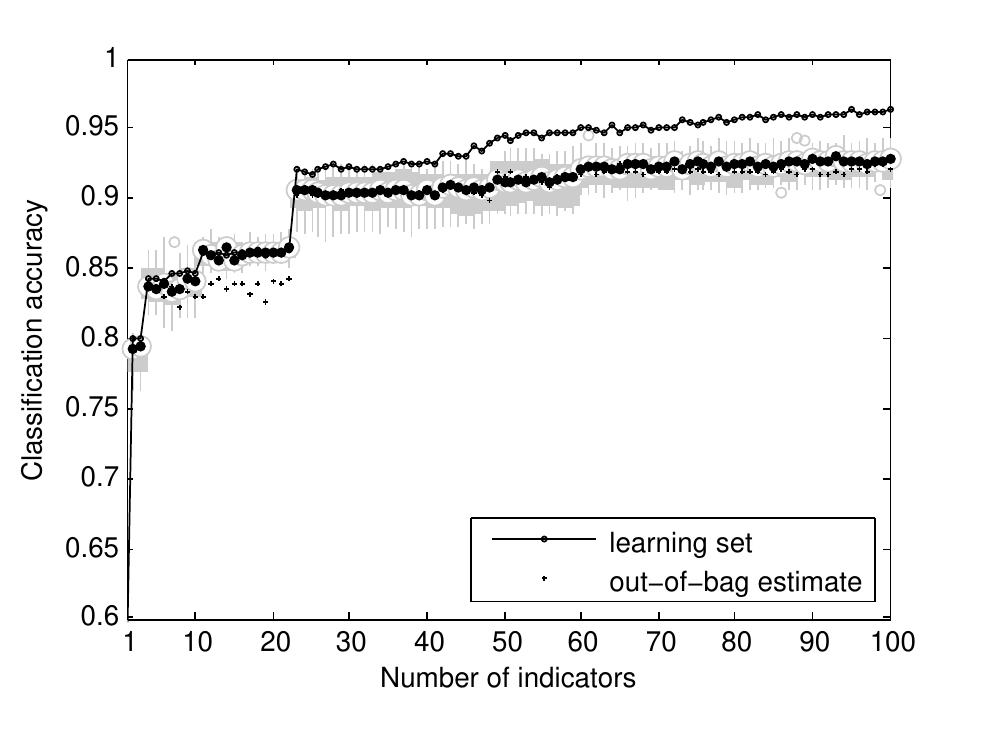}}
\caption{\textbf{Data set $A$ Random Forest}: classification accuracy on learning set (circle) as a function of the
  number of indicators. A boxplot gives the classification accuracies on the
  test subsets, summarized by its median (black dot inside a white
  circle). The estimation of those accuracies by the out-of-bag (oob) bootstrap
  estimate is shown by the crosses.}
\label{fig:rfboxsetA}
\end{figure}

\begin{figure}[htbp]
\centerline{\includegraphics[width=1\linewidth]{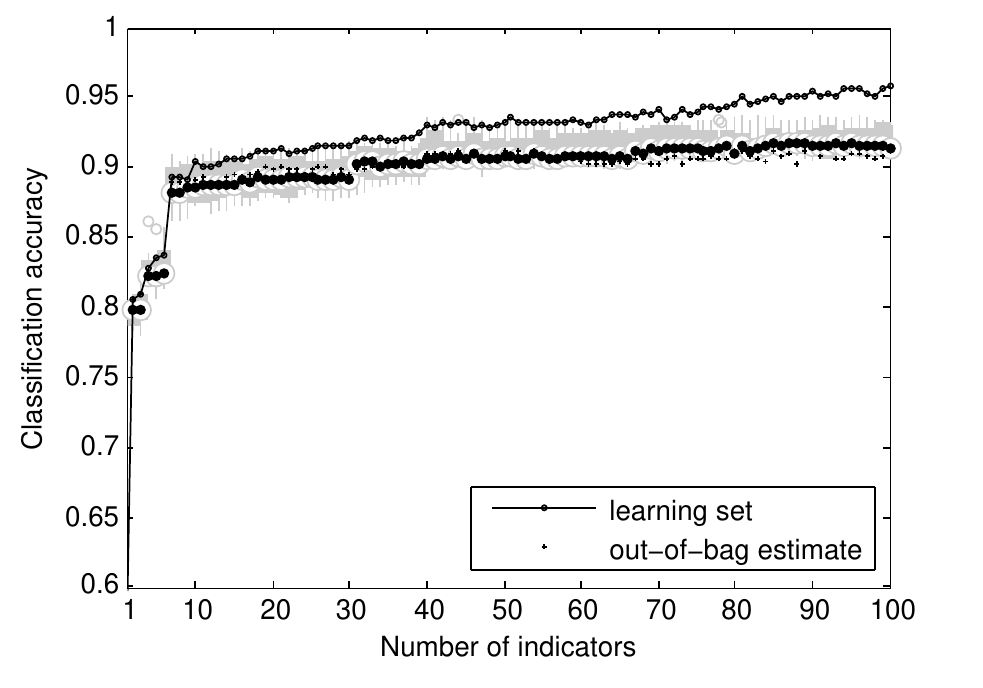}}
\caption{\textbf{Data set $B$ Random Forest}, see Figure \ref{fig:rfboxsetA}
  for details.}
\label{fig:rfboxsetD}
\end{figure}

\begin{figure}
\centering
\includegraphics[width=1\linewidth]{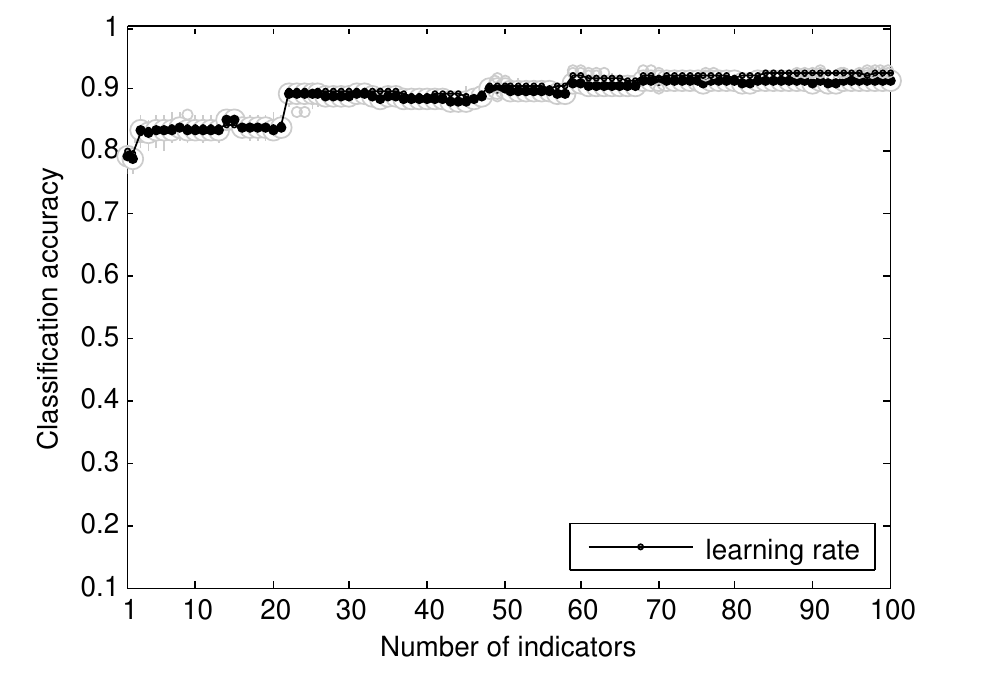}
\caption{\textbf{Data set $A$ Naive Bayes classifier}: classification accuracy on learning set (circle) as a function of the
  number of indicators. A boxplot gives the classification accuracies on the
  test subsets, summarized by its median (black dot inside a white
  circle). }
\label{fig:nbnboxsetA}
\end{figure}

\begin{figure}
\centering
\includegraphics[width=1\linewidth]{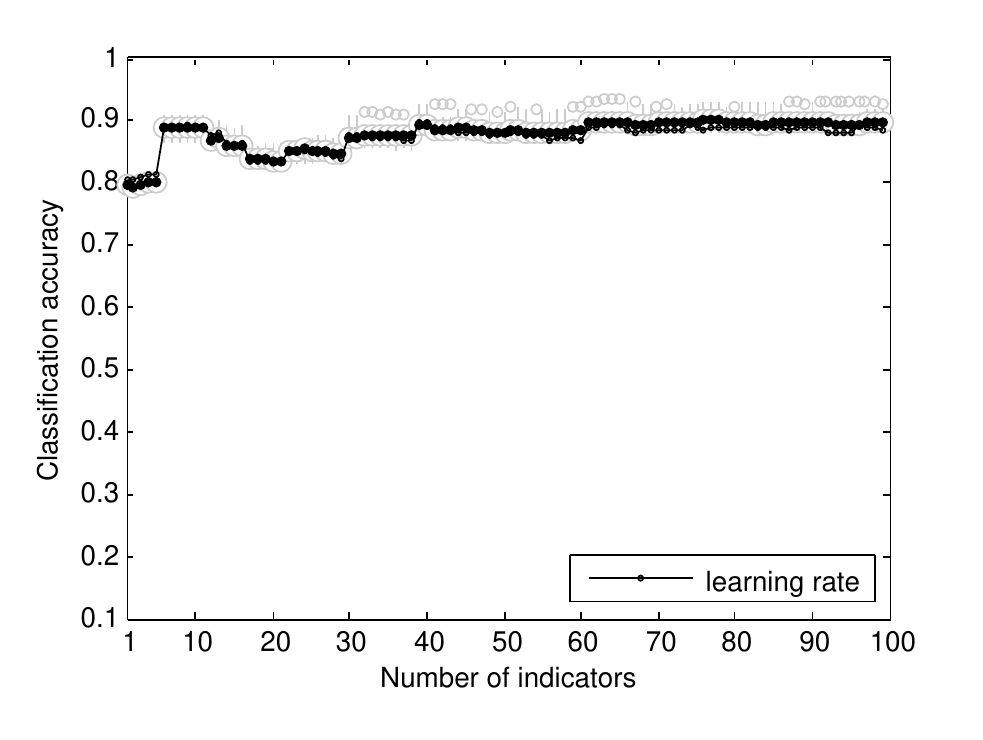}
\caption{\textbf{Data set $B$ Naive Bayes classifier}, see Figure
  \ref{fig:nbnboxsetA} for details.}
\label{fig:nbnboxsetD}
\end{figure}

Figures \ref{fig:rfboxsetA}, \ref{fig:rfboxsetD}, \ref{fig:nbnboxsetA} and
\ref{fig:nbnboxsetD} summarize the results for the 100 first indicators. The
classification accuracy of the Random Forest increases almost monotonously
with the number of indicators, but after roughly 25 to 30 indicators
(depending on the data set), performances on the test set tend to stagnate
(this is also the case of the out-of-bag estimate of the performances, which
shows, as expected, that the number of indicators could be selected using this
measure). In practice, this means that the proposed procedure can be used to
select the relevant indicators implementing this way an automatic tuning
procedure for the parameters of the expertly designed scores. 

Results for the Naive Bayes classifier are slightly more complex in the case
of the second data set, but they confirm that indicator selection is
possible. Moreover, reducing the number of indicators has here a very positive
effect on the classification accuracy of the Naive Bayes classifier which
reaches almost as good performances as the Random Forest. Notice that the
learning set performances of the Naive Bayes classifier are almost identical
to its test set performances (which exhibit almost no variability over the
slices of the full test set). This is natural because the classifier is based
on the estimation of the probability of observing a 1 value
\emph{independently} for each indicator, conditionally on the class. The
learning set contains at least 250 observations for each class, leading to a
very accurate estimation of those probabilities and thus to very stable
decisions. In practice one can therefore select the optimal number of
indicators using the learning set performances, without the need of a
cross-validation procedure.

\begin{figure}
\centering
\includegraphics[width=1\linewidth]{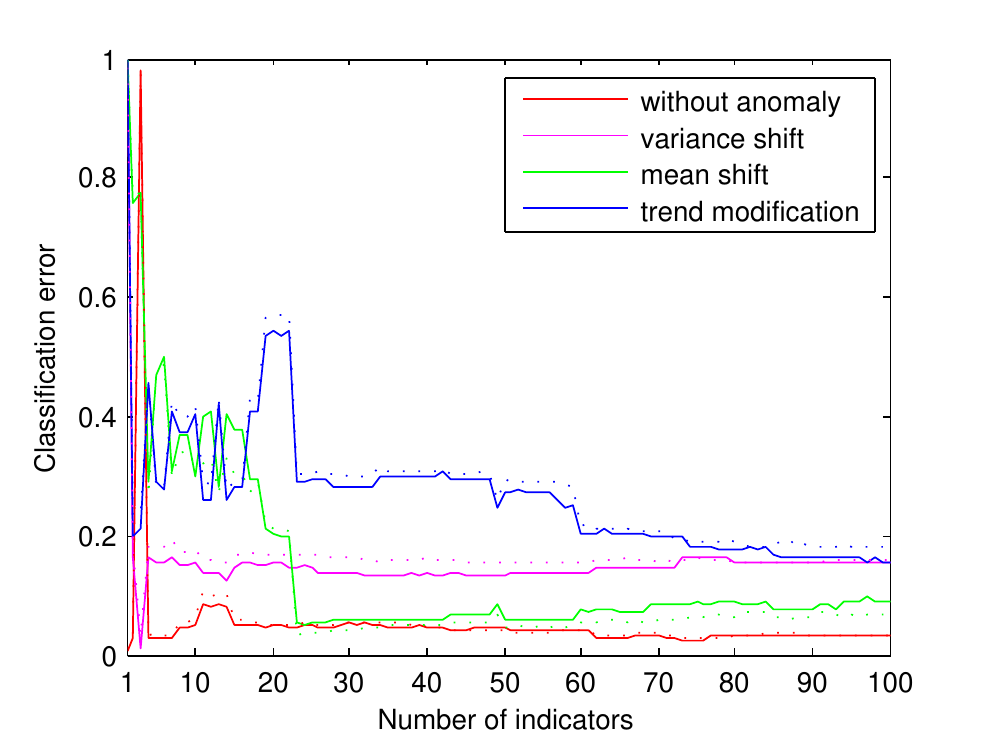}
\caption{\textbf{Data set $A$ Naive Bayes classifier}: classification error
  for each class on the training set (solid lines) and on the test set (dotted
  lines, average accuracies only).}
\label{fig:nbperclassA}
\end{figure}

It should be noted that significant jumps in performances can be observed in
all cases. This might be an indication that the ordering provided by the mRMR
procedure is not optimal. A possible solution to reach better indicator
subsets would be to use a wrapper approach, leveraging the computational
efficiency of both Random Forest and Naive Bayes construction. Meanwhile
Figure \ref{fig:nbperclassA} shows in more detail this phenomenon by
displaying the classification error class by class, as a function of the
number of indicators, in the case of data set $A$. The figure shows the
difficulty of discerning between mean shift and trend shift (for the latter,
no specific test have been included, on purpose). But as the strong decrease in
classification error when the 23-th indicator is added concerns both classes
(mean shift and trend shift), the ordering provided by mRMR could be
questioned. 

\subsection{Indicator selection}
Based on results shown on Figures \ref{fig:nbnboxsetA} and
\ref{fig:nbnboxsetD}, one can select an optimal number of binary indicators,
while enforcing a reasonable limit on this number to avoid flooding the human
operator with to many results. For instance Table \ref{tab:NB:selected} gives
the classification accuracy of the Naive Bayes classifier using the optimal
number of binary indicators between 1 and 30. 

\begin{table}[htbp]
\centering
\begin{tabular}{lcccc}
Data set & Training set acc. & Test set average acc. & \# of
indicators\\\hline
$A$ &0.8958 & 0.8911 (0.0125) & 23\\
$B$ &0.8828 & 0.8809 (0.0130) & 11\\ \hline
\end{tabular}
\caption{Classification accuracy of the Naive Bayesian network using the
optimal number binary indicators between 1 and 30. For the test set, we report the average
classification accuracy and its standard deviation between parenthesis.}
\label{tab:NB:selected}
\end{table}

While the performances are not as good as the ones of the Random Forest, they
are much improved compared to the ones reported in Table \ref{tab:fullNBN}. In
addition, the selected indicators can be shown to the human operator together
with the estimated probabilities of getting a positive result from each
indicator, conditionally on each class, shown on Table
\ref{tab:tenbest:A}. For instance here the first selected indicator,
$confu(2,3)$, is a confirmation indicator for the U test. It is positive when
there are 2 windows out of 3 consecutive ones on which a U test was
positive. The Naive Bayes classifier uses the estimated probabilities to reach
a decision: here the indicator is very unlikely to be positive if there is no
change or if the change is a variance shift. On the contrary, it is very
likely to be positive when there is a mean or a trend shift. While the table
does not ``explain'' the decisions made by the Naive Bayes classifier, it
gives easily interpretable hints to the human operator.  

\begin{table}[htbp]

\centering

\begin{tabular}{ccccc}

type of indicator& no change &variance&mean&trend\\\hline

confu(2,3)&0.010333&0.011&0.971&0.939\\

F test&0.020667&0.83&0.742&0.779\\

U test&0.027333&0.03&0.977&0.952\\

ratef(0.1)&0.0016667&0.69&0.518&0.221\\

confu(4,5)&0.034333&0.03&0.986&0.959\\

confu(3,5)&0.0013333&0.001&0.923&0.899\\

U test&0.02&0.022&0.968&0.941\\

F test&0.042&0.853&0.793&0.813\\

rateu(0.1)&0.00033333&0.001&0.906&0.896\\

confu(4,5)&0.019&0.02&0.946&0.927\\

conff(3,5)&0.052333&0.721&0.54&0.121\\

U test&0.037667&0.038&0.983&0.951\\

KS test&0.016&0.294&0.972&0.936\\

confu(3,5)&0.049&0.043&0.988&0.963\\

F test&0.030667&0.841&0.77&0.801\\

U test&0.043&0.043&0.981&0.963\\

lseqf(0.3)&0.0093333&0.749&0.59&0.36\\

rateu(0.1)&0.001&0.002&0.896&0.895\\

lsequ(0.1)&0.062667&0.06&0.992&0.949\\

confu(3,5)&0.025667&0.021&0.963&0.936\\

lseqf(0.3)&0.008&0.732&0.656&0.695\\

KS test&0.016333&0.088&0.955&0.93\\

confu(3,5)&0&0&0.003&0.673\\\hline

\end{tabular}

\caption{The 23 best indicators according to mRMR for data set
$A$. Confu(k,n) corresponds to a positive Mann–Whitney–Wilcoxon U test on
k windows out of n consecutive ones. Conff(k,n) is the same thing for the
F-test. Ratef($\alpha$) corresponds to a positive F-test on $\alpha\times m$
windows out of $m$. Lseqf($\alpha$) corresponds to a positive F-test on
$\alpha\times m$ consecutive windows out of $m$. Lsequ($\alpha$) is the same for a U test.
Detailed parameters of the indicators have been omitted for brevity.}

\label{tab:tenbest:A}
\end{table}

\section{Conclusion and perspectives}
This paper proposes a general methodology that combines expert knowledge with
feature selection and automatic classification to design accurate anomaly
detector and classifier. The main idea is to build from expert knowledge
parametric anomaly scores associated to range of plausible parameters. From
those scores, hundreds of binary indicators are generated in a way that covers
the parameter space as well as introduce simple confirmation indicators.  This
turns anomaly detection into a classification problem with a very high number
of binary features. Using a feature selection technique, one can reduce the
number of useful indicators to a humanly manageable number. This allows a
human operator to understand at least partially how a decision is reached by
an automatic classifier. This is favored by the choice of the indicators which
are based on expert knowledge. A very interesting byproduct of the
methodology is that it can work on very different original data as long as
expert decision can be modeled by a set of parametric anomaly scores. This was
illustrated by working on signals of different lengths. 

The methodology has been shown sound using simulated data. Using a reference
high performance classifier, Random Forests, the indicator generation
technique covers sufficiently the parameter space to obtain high
classification rate. Then, the feature selection mechanism (here a simple
forward technique based on mRMR) leads to a reduced number of indicators (23
for one of the data set) with good predictive performances when paired with a
simpler classifier, the Naive Bayes classifier. As shown in the experiments,
the class conditional probabilities of obtaining a positive value for those
indicators provide interesting insights on the way the Naive Bayes classifier
takes a decision. 

In order to justify the costs of collecting a sufficiently large real world
labelled data set in our context (engine health monitoring), additional
experiments are needed. In particular, multivariate data must be studied in
order to simulate the case of a complex system made of numerous
sub-systems. This will naturally lead to more complex anomaly models. We also
observed possible limitations of the feature selection strategy used here as
the performances displayed abrupt changes during the forward procedure. More
computationally demanding solutions, namely wrapper ones, will be studied to
confirm this point. 

It is also important to notice that the classification accuracy is not the
best way of evaluating the performances of a classifier in the health
monitoring context. Firstly, health monitoring involves intrinsically a strong
class imbalance \cite{japkowicz2002class}. Secondly, health monitoring is a
cost sensitive area because of the strong impact on airline profit of an
unscheduled maintenance. It is therefore important to take into account
specific asymmetric misclassification cost to get a proper performance
evaluation.

% ****************************************************************************
% BIBLIOGRAPHY AREA
% ****************************************************************************

\bibliography{Biblio}

% Generated by IEEEtran.bst, version: 1.13 (2008/09/30)
\begin{thebibliography}{10}
\providecommand{\url}[1]{#1}
\csname url@samestyle\endcsname
\providecommand{\newblock}{\relax}
\providecommand{\bibinfo}[2]{#2}
\providecommand{\BIBentrySTDinterwordspacing}{\spaceskip=0pt\relax}
\providecommand{\BIBentryALTinterwordstretchfactor}{4}
\providecommand{\BIBentryALTinterwordspacing}{\spaceskip=\fontdimen2\font plus
\BIBentryALTinterwordstretchfactor\fontdimen3\font minus
  \fontdimen4\font\relax}
\providecommand{\BIBforeignlanguage}[2]{{%
\expandafter\ifx\csname l@#1\endcsname\relax
\typeout{** WARNING: IEEEtran.bst: No hyphenation pattern has been}%
\typeout{** loaded for the language `#1'. Using the pattern for}%
\typeout{** the default language instead.}%
\else
\language=\csname l@#1\endcsname
\fi
#2}}
\providecommand{\BIBdecl}{\relax}
\BIBdecl

\bibitem{chandola2009anomaly}
V.~Chandola, A.~Banerjee, and V.~Kumar, ``Anomaly detection: A survey,''
  \emph{ACM Computing Surveys (CSUR)}, vol.~41, no.~3, p.~15, 2009.

\bibitem{basseville1995detection}
M.~Basseville and I.~V. Nikiforov, ``Detection of abrupt changes: theory and
  applications,'' \emph{Journal of the Royal Statistical Society-Series A
  Statistics in Society}, vol. 158, no.~1, p. 185, 1995.

\bibitem{fleuret-2004}
F.~Fleuret, ``Fast binary feature selection with conditional mutual
  information,'' \emph{Journal of Machine Learning Research (JMLR)}, vol.~5,
  pp. 1531--1555, 2004.

\bibitem{hegedus2011methodology}
J.~Hegedus, Y.~Miche, A.~Ilin, and A.~Lendasse, ``Methodology for
  behavioral-based malware analysis and detection using random projections and
  k-nearest neighbors classifiers,'' in \emph{Computational Intelligence and
  Security (CIS), 2011 Seventh International Conference on}.\hskip 1em plus
  0.5em minus 0.4em\relax IEEE, 2011, pp. 1016--1023.

\bibitem{rabenoroinstants}
T.~Rabenoro and J.~Lacaille, ``Instants extraction for aircraft engine
  monitoring,'' \emph{AIAA Infotech@Aerospace}, 2013.

\bibitem{come2010aircraft}
E.~C{\^o}me, M.~Cottrell, M.~Verleysen, and J.~Lacaille, ``Aircraft engine
  health monitoring using self-organizing maps,'' in \emph{Advances in Data
  Mining. Applications and Theoretical Aspects}.\hskip 1em plus 0.5em minus
  0.4em\relax Springer, 2010, pp. 405--417.

\bibitem{flandrois2009expertise}
X.~Flandrois, J.~Lacaille, J.-R. Masse, and A.~Ausloos, ``Expertise transfer
  and automatic failure classification for the engine start capability
  system,'' \emph{AIAA Infotech, Seattle, WA}, 2009.

\bibitem{lacaille2009maturation}
J.~Lacaille, ``A maturation environment to develop and manage health monitoring
  algorithms,'' \emph{PHM, San Diego, CA}, 2009.

\bibitem{kotsiantis2007supervised}
S.~B. Kotsiantis, I.~Zaharakis, and P.~Pintelas, ``Supervised machine learning:
  A review of classification techniques,'' 2007.

\bibitem{breiman2001random}
L.~Breiman, ``Random forests,'' \emph{Machine learning}, vol.~45, no.~1, pp.
  5--32, 2001.

\bibitem{breiman1984classification}
L.~Breiman, J.~H. Friedman, R.~A. Olshen, and C.~J. Stone, ``Classification and
  regression trees. wadsworth \& brooks,'' \emph{Monterey, CA}, 1984.

\bibitem{koller2009probabilistic}
D.~Koller and N.~Friedman, \emph{Probabilistic graphical models: principles and
  techniques}.\hskip 1em plus 0.5em minus 0.4em\relax The MIT Press, 2009.

\bibitem{guyon2003introduction}
I.~Guyon and A.~Elisseeff, ``An introduction to variable and feature
  selection,'' \emph{The Journal of Machine Learning Research}, vol.~3, pp.
  1157--1182, 2003.

\bibitem{peng2005feature}
H.~Peng, F.~Long, and C.~Ding, ``Feature selection based on mutual information
  criteria of max-dependency, max-relevance, and min-redundancy,''
  \emph{Pattern Analysis and Machine Intelligence, IEEE Transactions on},
  vol.~27, no.~8, pp. 1226--1238, 2005.

\bibitem{japkowicz2002class}
N.~Japkowicz and S.~Stephen, ``The class imbalance problem: A systematic
  study,'' \emph{Intelligent data analysis}, vol.~6, no.~5, pp. 429--449, 2002.

\end{thebibliography}

\bibliographystyle{IEEEtran}
% ****************************************************************************
% END OF BIBLIOGRAPHY AREA
% ****************************************************************************

% trigger a \newpage just before the given reference
% number - used to balance the columns on the last page
% adjust value as needed - may need to be readjusted if
% the document is modified later
%\IEEEtriggeratref{8}
% The "triggered" command can be changed if desired:
%\IEEEtriggercmd{\enlargethispage{-5in}}

% references section
% NOTE: BibTeX documentation can be easily obtained at:
% http://www.ctan.org/tex-archive/biblio/bibtex/contrib/doc/

% can use a bibliography generated by BibTeX as a .bbl file
% standard IEEE bibliography style from:
% http://www.ctan.org/tex-archive/macros/latex/contrib/supported/IEEEtran/testflow/bibtex
%\bibliographystyle{IEEEtran.bst}
% argument is your BibTeX string definitions and bibliography database(s)
%\bibliography{IEEEabrv,../bib/paper}
%
% <OR> manually copy in the resultant .bbl file
% set second argument of \begin to the number of references
% (used to reserve space for the reference number labels box)

\def\V{\rm vol.~}
\def\N{no.~}
\def\pp{pp.~}
\def\Pot{\it Proc. }
\def\IJCNN{\it International Joint Conference on Neural Networks\rm }
\def\ACC{\it American Control Conference\rm }
\def\SMC{\it IEEE Trans. Systems\rm , \it Man\rm , and \it Cybernetics\rm }

\def\handb{ \it Handbook of Intelligent Control: Neural\rm , \it
    Fuzzy\rm , \it and Adaptive Approaches \rm }

\end{document}